\documentclass[]{informs3}

\DoubleSpacedXII

\usepackage{natbib}
\bibpunct[, ]{(}{)}{,}{a}{}{,}%
\def\bibfont{\small}%
\TheoremsNumberedThrough     

\EquationsNumberedThrough    

\MANUSCRIPTNO{}

\usepackage{algorithm}
\usepackage{algcompatible}
\usepackage{algpseudocode}

\usepackage{booktabs}       
\usepackage{comment}
\usepackage{graphicx}
\usepackage[font=onehalfspacing]{quoting}

\usepackage{pdflscape}

\usepackage{multicol}
\usepackage{epigraph}

\usepackage{subcaption}
\usepackage{tikz}
\usepackage{varwidth} 
\usepackage{wrapfig}
\usetikzlibrary{arrows,arrows.meta,chains,er,fit,positioning,shapes,tikzmark,shapes.geometric} 

\usepackage{stringstrings} 
\usepackage{color} 
\usepackage[stable]{footmisc}

\makeatletter 
\let\OldStatex\Statex
\renewcommand{\Statex}[1][0]{%
	\setlength\@tempdima{\algorithmicindent}%
	\OldStatex\hskip\dimexpr#1\@tempdima\relax}
\let\OldState\State
\renewcommand{\State}[1][0]{%
	\setlength\@tempdima{\algorithmicindent}%
	\OldState\hskip\dimexpr#1\@tempdima\relax}

\newcommand{\leqnomode}{\tagsleft@true}
\newcommand{\reqnomode}{\tagsleft@false}

\makeatother 

\newcommand\newsubcap[1]{\phantomcaption%
       \caption*{\footnotesize\figurename~\thefigure\thesubfigure: #1}}

\newcounter{tmkcount}
\tikzset{%
	tikzmark suffix={-\thetmkcount},%
	defaultCodeBox/.style={draw=black}%
}

\newcommand{\drawCodeBox}[4]{%
	\begin{tikzpicture}[remember picture,overlay]
	\coordinate (start) at ([yshift=1.0ex]pic cs:#2);
	\coordinate (middle) at (pic cs:#3);
	\coordinate (end) at ([yshift=-0.2ex]pic cs:#4);
	\node[inner sep=2pt,#1,fit=(start) (middle) (end)] {};
	\end{tikzpicture}%
}

\newcommand{\BeginBox}[1]{%
	\drawCodeBox{#1}{beginCB}{middleCB}{endCB}%
	\tikzmark{beginCB}\tikzmark{middleCB}%
}

\algdef{S}[WHILE]{BoxedWhile}[2][defaultCodeBox]{\BeginBox{#1}\algorithmicwhile\ #2\ \algorithmicdo}%
\algdef{S}[FOR]{BoxedFor}[2][defaultCodeBox]{\BeginBox{#1}\algorithmicfor\ #2\ \algorithmicdo}%
\algdef{S}[FOR]{BoxedForAll}[2][defaultCodeBox]{\BeginBox{#1}\algorithmicforall\ #2\ \algorithmicdo}%
\algdef{S}[LOOP]{BoxedLoop}[1][defaultCodeBox]{\BeginBox{#1}\algorithmicloop}%
\algdef{S}[REPEAT]{BoxedRepeat}[1][defaultCodeBox]{\BeginBox{#1}\algorithmicrepeat}%
\algdef{S}[IF]{BoxedIf}[2][defaultCodeBox]{\BeginBox{#1}\algorithmicif\ #2\ \algorithmicthen}%
\algdef{C}[IF]{IF}{BoxedElsIf}[2][defaultCodeBox]{\BeginBox{#1}\algorithmicelse\ \algorithmicif\ #2\ \algorithmicthen}%
\algdef{Ce}[ELSE]{IF}{BoxedElse}{EndIf}[1][defaultCodeBox]{\BeginBox{#1}\algorithmicelse}%

\AtBeginDocument{%
  \providecommand\BibTeX{{%
    \normalfont B\kern-0.5em{\scshape i\kern-0.25em b}\kern-0.8em\TeX}}}

\theoremstyle{remark}
\newtheorem{property}{Property}

\newcommand{\useritembias}{user-item bias}
\newcommand{\uib}{UIB}
\newcommand{\useruserknn}{user-user $k$-nearest neighbors}
\newcommand{\uuknn}{KNN} 

\newcommand{\graphname}{operation context}
\newcommand{\Graphname}{Operation Context}

\newcommand{\frameworkname}{measuring algorithmic interpretability}
\newcommand{\Frameworkname}{Measuring algorithmic interpretability}
\newcommand{\scorename}{cognitive complexity}
\newcommand{\Scorename}{Cognitive complexity}

\renewcommand{\theARTICLETOP}{}

\begin{document}
	\RUNAUTHOR{Lalor and Guo}
	\RUNTITLE{\capitalizetitle{\frameworkname}}
	
	\TITLE{Measuring algorithmic interpretability:\\
	A human-learning-based framework and\\
	the corresponding cognitive complexity score}
	\ARTICLEAUTHORS{%
		\AUTHOR{John P. Lalor, Hong Guo}
		\AFF{IT, Analytics, and Operations Department\\
                  Mendoza College of Business\\
                  University of Notre Dame, Notre Dame, IN 46556 \EMAIL{john.lalor@nd.edu}, \EMAIL{hguo@nd.edu}} 
	} 

	\ABSTRACT{
		Algorithmic interpretability is necessary to build trust, ensure fairness, and track accountability. 
		However, there is no existing formal measurement method for algorithmic interpretability. 
		In this work, we build upon programming language theory and cognitive load theory to develop a framework for measuring algorithmic interpretability.
		The proposed measurement framework reflects the process of a human learning an algorithm.
		We show that the measurement framework and the resulting cognitive complexity score have the following desirable properties – \textit{universality}, \textit{computability}, \textit{uniqueness}, and \textit{monotonicity}.
		We illustrate the measurement framework through a toy example, describe the framework and its conceptual underpinnings, and demonstrate the benefits of the framework, in particular for managers considering tradeoffs when selecting algorithms.
	}

	\KEYWORDS{algorithmic interpretability, human learning, cognitive complexity, programming language theory, cognitive load theory} 
	\newcommand{\suma}{\Large$+$}
	\newcommand{\mina}{\Large$-$}
	\newcommand{\diva}{\Large$\div$}
	\newcommand{\hl}[2]{{\color{#1}\bfseries [[#2]]}} 
	\newcommand{\todo}[1]{\hl{red}{#1}}
	\newcommand{\rewrite}[1]{\hl{green}{#1}}
	\maketitle
	
	\leqnomode 

	\tikzset{%
		data/.style    = {draw, thick, circle, minimum height = 1em,
		minimum width = 1em},
		weight/.style    = {draw, thick, circle, minimum height = 1em,
		minimum width = 1em},
		block/.style    = {draw, thick, circle, minimum height = 2em,
			minimum width = 1.5em, execute at begin node={\begin{varwidth}{7em}},
				execute at end node={\end{varwidth}}},
		blockCFG/.style    = {draw, thick, minimum height = 2em,
			minimum width = 1.5em, execute at begin node={\begin{varwidth}{7em}},
				execute at end node={\end{varwidth}}},
		controlBlock/.style    = {draw, thick, diamond, minimum height = 2em,
			minimum width = 1.5em, execute at begin node={\begin{varwidth}{7em}},
				execute at end node={\end{varwidth}}},
		outer/.style={draw, rectangle,inner sep=5pt},
		endpoint/.style      = {draw, diamond, dashed, node distance = 0.5cm}, 
		-|/.style={to path={-| (\tikztotarget)}},
		|-/.style={to path={|- (\tikztotarget)}},
		element/.style args={#1:#2:#3}{name=#3,%
			draw, fill=white,
			text width=3mm, minimum height=5mm, outer sep=0pt,
			node contents={},
			append after command={node[fill=green!30,
				minimum size=2mm, inner sep=2pt, outer sep=0.4pt,
				anchor=south, font=\tiny ]
				at (#3.south) {#1}},
			append after command={node[text width=5mm-3mm,
				inner sep=1pt, outer sep=2pt,
				align=center, font=\small, anchor=north]
				at (#3.north) {#2}}
		},
	}


\section{Introduction}

\setlength\epigraphrule{0pt}
\setlength{\epigraphwidth}{0.42\textwidth}
\epigraph{If you can't measure it, you can't improve it.}{---Peter Drucker (apocryphal)}

Algorithms built on artificial intelligence (AI) and machine learning technologies are being deployed in a wide variety of areas such as healthcare, criminal justice, and financial services \citep{lucasImpactfulResearchTransformational2013,chouldechovaFairPredictionDisparate2017,de2019bias,mckinneyInternationalEvaluationAI2020,stricklandOpenAIGPT3Speaks2021}.
With this widespread deployment has also come calls for these algorithms to be more interpretable in order to build trust, ensure fairness, and track accountability \citep{hardtEqualityOpportunitySupervised2016a,xiaoOpportunitiesChallengesDeveloping2018a,kearnsEthicalAlgorithmScience2019,mittelstadtExplainingExplanationsAI2019a,zhangEffectConfidenceExplanation2020}.
Such calls have even risen to the level of legislation, for example the European Union's General Data Protection Regulation (GDPR) requiring ``meaningful information about the logic involved'' in algorithmic decision making.


Despite these concerns across a wide range of stakeholders including managers, users, policymakers, learners, and educators, there is presently no formal measurement method for algorithmic interpretability \citep{liInterpretableDeepLearning2021}.
Measurement ``is the assignment of numbers to objects or events according to rule''~\citep{stevensMeasurementMan1958}.
Algorithmic interpretability refers to the understandability and sensemaking for a particular human with respect to a specific algorithm~\citep{berenteManagingArtificialIntelligence2021}.\footnote{In the existing literature, the concept of algorithmic interpretability is often intertwined with concepts such as transparency and explainability. We refer readers to \cite{berenteManagingArtificialIntelligence2021} for a detailed discussion of the concepts and their similarities and differences.}
Interpretability is often assumed based on intuitive understanding of algorithms or estimated based on model characteristics (e.g., tree depth or weight sparsity) or outputs of downstream user studies.\footnote{Some researchers argue that inherently interpretable models are the way forward, as interpretation methods for the output of so-called ``black-box'' models generate interpretations that are not faithful to what the original model computes and can potentially mislead \citep{liptonMythosModelInterpretability2018a,rudinStopExplainingBlack2019a}.}
Even the GDPR regulations do not specify the level of detail required to meet the standard of ``meaningful information'' and do not account for different individuals' expectations for the same algorithm~\citep{goodmanEuropeanUnionRegulations2017a}.
Without a formal framework, we can't measure, compare, or manage algorithmic interpretability.

In this work we build upon programming language theory and cognitive load theory to develop a framework for measuring algorithmic interpretability.
The proposed measurement framework reflects the process of a human learning an algorithm.
We show that the measurement framework and the resulting cognitive complexity score have the following desirable properties – \textit{universality}, \textit{computability}, \textit{uniqueness}, and \textit{monotonicity}.
As our opening quote indicates, having a framework for measuring interpretability sets a foundation for improving algorithmic interpretability across algorithms.

\begin{figure}[h!]
	\centering 
	\includegraphics[width=0.75\columnwidth]{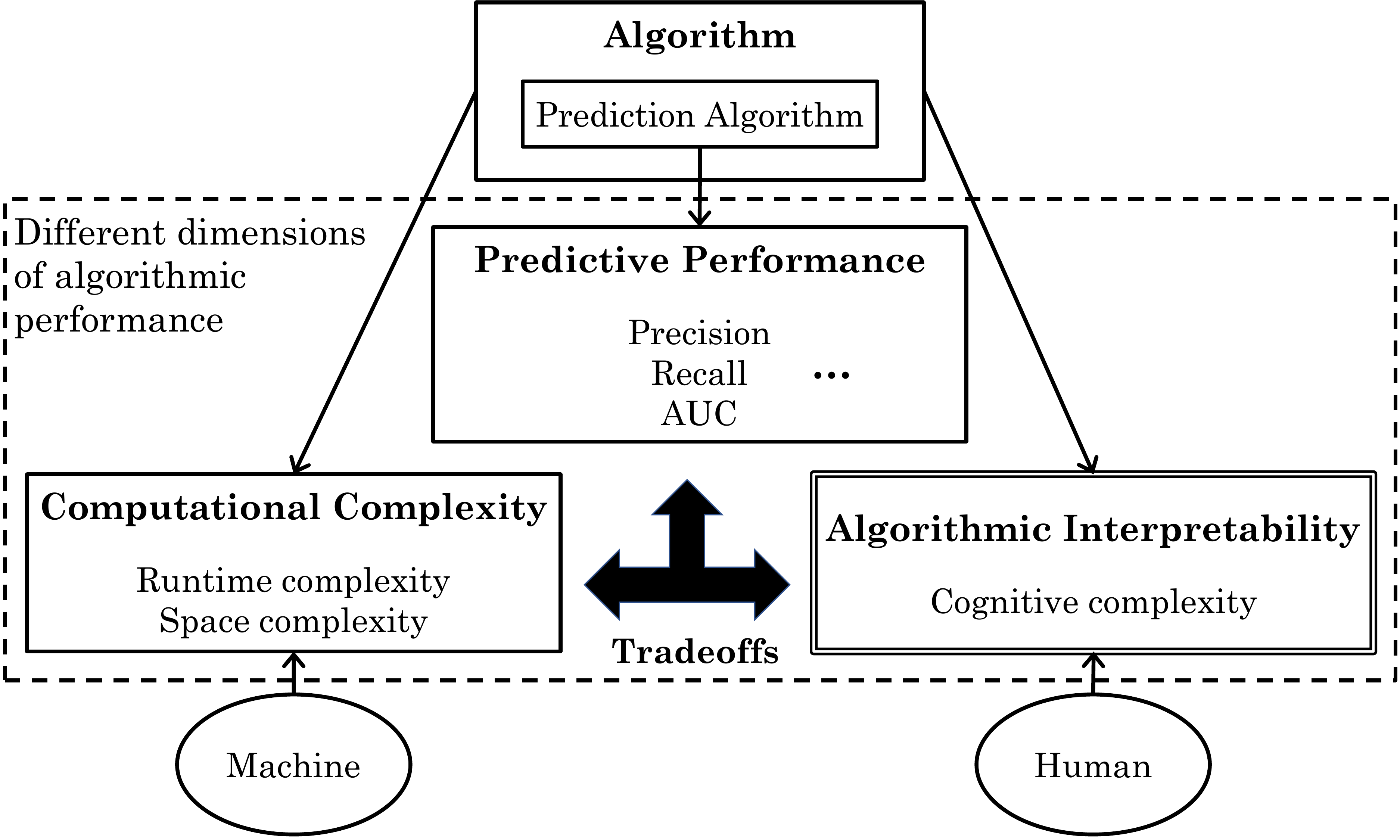}
	\caption{Algorithmic interpretability: A new dimension of algorithm performance}
	\label{fig:position}
\end{figure}

\Frameworkname~bridges the gap between human and machine evaluation (Figure \ref{fig:position}).
Computational complexity (such as runtime complexity and space complexity) measure an algorithm's mechanical, machine-oriented nature.\footnote{Here, ``machine-oriented'' 
	should not be confused with ``machine-dependent.'' 
	Computational complexity is not machine-dependent, i.e., it is the same for a state-of-the-art machine and an old computer. 
	Computational complexity is machine-oriented, measuring how difficult it is for computers to run the algorithm. }
Predictive algorithms are also measured by their performance on specific tasks, 
e.g., accuracy or precision/recall/F-score metrics.
\Scorename~represents the intrinsic, human-oriented difficulty of understanding 
an algorithm.
When choosing an algorithm for a task and group of users, tradeoffs among all performance dimensions must be carefully evaluated.
In this work (indicated by the double-lined rectangle in Figure 
\ref{fig:position}), we measure this as yet unexplored dimension of algorithmic 
performance -- \scorename.

The rest of the paper is organized as follows.
In Section \ref{sec:relatedwork}, we describe related work on interpretability.
In Section \ref{sec:theory}, we discuss the necessary theoretical background in programming language theory from computer science and cognitive load theory from cognitive science.
In Section \ref{sec:toyexample}, we walk through a toy example to detail the specifics of our framework.
In Section \ref{sec:framework}, we describe the details of our framework.
In Section \ref{sec:properties}, we discuss the important properties of the framework.
In Section \ref{sec:recsys}, we show a detailed real-world business example of the framework as applied to recommender systems, which are widely used in a number of business contexts~\citep[e.g.,][]{lee2021product}.
In Section \ref{sec:conclusion}, we conclude with a detailed discussion of our contributions to research, including promising areas for future work (\S \ref{ssec:cResearch}), and contributions to practice for specific stakeholders such as managers, users, policymakers, learners, and educators (\S \ref{ssec:cPractice}).
To offer a consolidated point of reference for readers, in Appendix \ref{sec:constructs}, we summarize the key concepts from the theories we build upon (programming language theory and cognitive load theory).

\section{Related Work}
\label{sec:relatedwork}

In this section, we first discuss the prior research on algorithmic interpretability broadly, then focus on the question of how algorithmic interpretability is measured.

\subsection{Algorithmic Interpretability}

Algorithmic interpretability refers to the understandability and sensemaking for a particular human with respect to a specific algorithm \citep{berenteManagingArtificialIntelligence2021}.
Put another way, ``for a system to be interpretable, it must produce descriptions that are simple enough for a person to understand using a vocabulary that is meaningful to the user'' \citep{gilpinExplainingExplanationsOverview2018a}.\footnote{For a discussion on how algorithmic interpretability relates to concepts such as explainable AI (xAI) and algorithmic transparency, we refer the reader to existing surveys and overviews on the topic \citep{doshi-velezRigorousScienceInterpretable2017b,adadiPeekingBlackBoxSurvey2018a,liptonMythosModelInterpretability2018a,murdochDefinitionsMethodsApplications2019a}.}

During the development of new interpretable algorithms, there are three key considerations concerning what a human is meant to interpret and how he or she should do so.
The first dimension involves that which is being interpreted: the algorithm or the algorithm's output.
This dimension of interpretability can be categorized into \textit{intrinsic interpretability} or \textit{post-hoc interpretability}
\citep{liptonMythosModelInterpretability2018a,murdochDefinitionsMethodsApplications2019a}.
The target for intrinsic interpretability is the algorithm itself, while the target for post-hoc interpretability is the output (e.g., a model prediction).
Most existing methods provide post-hoc interpretability \citep[e.g.,][]{abdulTrendsTrajectoriesExplainable2018b,rossEvaluatingInterpretabilityGenerative2021b}.

The second dimension covers the scope of the interpretation, and can be categorized into \textit{global} interpretability versus \textit{local} interpretability.
Global interpretability methods describe model behavior in the average case and therefore cannot be example-specific \citep{lageHumanintheLoopInterpretabilityPrior2018a}.
Local interpretability methods allow for interpreting the outputs of a model for specific inputs \citep{poursabzi-sangdehManipulatingMeasuringModel2021b}.

The last dimension involves the generality of the method.
\textit{Universal} methods are applicable to all algorithms \citep[e.g.,][provide a universal framework for measurement, but it requires model-specific proxies for use.]{lageHumanintheLoopInterpretabilityPrior2018a}.
\textit{Model-specific} methods only apply to specific algorithms, typically because of the inputs required for calculation \citep[e.g., linear model sparsity as a quantification of interpretability,][]{abdulUppercaseCOGAMMeasuringModerating2020}.

\subsection{Measuring Algorithmic Interpretability}

In recent years, there has been a swell of research on making algorithms more interpretable.
The vast majority of work covers designing ``more interpretable'' algorithms \citep[][]{doshi-velezRigorousScienceInterpretable2017b,adadiPeekingBlackBoxSurvey2018a,liptonMythosModelInterpretability2018a,murdochDefinitionsMethodsApplications2019a}.
However, there is little consensus on how to measure algorithmic interpretability.

\cite{murdochDefinitionsMethodsApplications2019a} propose five types of model-based interpretability (sparsity, simulatability, modularity, domain-based feature engineering, and model-based feature engineering) but do not include any information on how to measure each type.
Similarly, Axis 2 in the framework of \cite{adadiPeekingBlackBoxSurvey2018a} covers ``XAI Measurement'' but it only considers the evaluation of output explanations and not of the interpretability of the models themselves.\footnote{In addition, works cited by \cite{adadiPeekingBlackBoxSurvey2018a} consider downstream explanations, and do not include measurements of the algorithm.}

There is presently no formal framework for the rigorous measurement of algorithmic interpretability.
The existing work either uses intuition-driven proxies or user studies to quantify interpretability, but does not formally measure the interpretability of an algorithm (see Table \ref{tab:litreview} for a summary of related studies).
\cite{abdulUppercaseCOGAMMeasuringModerating2020} count the visual chunks in line chart repesentations of models to estimate the visual load of processing the explanation.
\cite{rossEvaluatingInterpretabilityGenerative2021b} operationalize understandability as the ability for humans to reconstruct input images based on different dimensionality reduction techniques.
\cite{lageHumanintheLoopInterpretabilityPrior2018a} take an optimization approach to include response time from user study outputs in an estimation task for identifying an interpretable model.
\cite{poursabzi-sangdehManipulatingMeasuringModel2021b} conduct extensive user studies to assess how model changes affect how well a human can predict a model's output.

\newcolumntype{C}[1]{>{\centering\let\newline\\\arraybackslash\hspace{0pt}}m{#1}}

\begin{table}[h] \centering \footnotesize
  \caption{Review of Existing Literature on Measuring Algorithmic Interpretability}
  \label{tab:litreview}
  \begin{tabular}{m{0.2\linewidth}C{0.15\linewidth}C{0.2\linewidth}C{0.15\linewidth}C{0.15\linewidth}}
    \toprule 
    Paper & Measurement \newline Type & Measure &  Intrinsic \text{or post-hoc} & Universal \text{or specific} \\
    \midrule
    \cite{lageHumanintheLoopInterpretabilityPrior2018a} & Reflective & Simulatability speed &  Post-hoc&Universal \\
    \cite{abdulUppercaseCOGAMMeasuringModerating2020} & Reflective & Simulatability speed & Post-hoc & Specific \\
    \cite{poursabzi-sangdehManipulatingMeasuringModel2021b} & Reflective & Simulatability accuracy &  Intrinsic and Post-hoc & Specific \\
    \cite{rossEvaluatingInterpretabilityGenerative2021b} & Reflective & Reconstruction accuracy &  Post-hoc & Specific \\
    This paper & Formative & Cognitive complexity &  Intrinsic & Universal \\
    \bottomrule
  \end{tabular}
\end{table}

Intrinsic interpretability is itself a latent construct.
To measure it, we rely on observable values.
It is important to consider the relationship between the observed values and the latent construct.
If a change in the latent variable is likely to change the measure, then the measure is \textit{reflective}.
For example, replacing a decision tree with a neural network presumably decreases the latent interpretability, which leads to changes in downstream observed values (such as longer simulation speed times or lower reported understanding).
However, if a change in the measure causes changes in the latent variable, then the measure is \textit{formative}.
Using a decision tree as an example, if the measure of interpretability is tree depth, changing the depth of the tree will change the latent interpretability.
Most measures used in xAI research are reflective (e.g., simulatability, output prediction, accuracy, etc.).
If you change the latent variable (intrinsic interpretability), then that will affect the measures.
Our cognitive complexity, on the other hand, is a formative measure.
As you modify an algorithm to change its cognitive complexity score or change the user group, this will affect the interpretability.
In other words, our cognitive complexity measure is determined by the algorithm and the user group involved.

At present, most work on measuring interpretability relies on user studies.
For a given algorithm, a user study is designed to measure some human-related construct (e.g., simulation speed, reported ease of use, or human accuracy on a downstream task).
These results give insights into how users perceive algorithms, or how algorithms affect user performance and behavior.
This approach has a number of benefits.
As humans are central to the question of algorithmic interpretability, it makes sense for human evaluation with user studies to be a key consideration for system design and tracking improvements in the field \citep{peffersDesignScienceResearch2007a}.
User studies where a system is evaluated in the task it was designed for by the true end-users allow researchers to ensure ``that the system delivers on its intended task'' \citep{doshi-velezRigorousScienceInterpretable2017b}.
However, user studies also have several downsides. A salient downside of user studies is the high cost of recruiting and engaging subjects.
In addition, such a data-driven approach in user studies cannot answer the question of why one algorithm is more interpretable than another.
What's more, user-studies are inherently ``task-driven'' evaluations \citep{doshi-velezRigorousScienceInterpretable2017b}.

On the other hand, a theory-driven approach can address these downsides.
First, a model of interpretability could answer the ``why'' question.
Instead of using humans to provide the \textit{measure} (in the form of user study data), a theory-driven approach seeks to \textit{model} human behavior or other characteristics (such as expertise).
This type of theory-driven approach complements the existing data-driven approach in user studies, and is the focus of our work.
Unlike the existing literature on algorithmic interpretability, we directly model the human learning process.
Our framework does not require user studies, which is beneficial for two reasons.
First, any algorithm and any user group can be operationalized according to the framework, which gives it broader applicability than work driven by specific user studies.
Second, the framework is automatically computable, making it more cost-effective than user studies.

\section{Theoretical Background}
\label{sec:theory}

Our framework for measuring algorithmic interpretability builds upon programming language theory from computer science and cognitive load theory from cognitive science.\footnote{We provide a table of relevant concepts and descriptions in Appendix \ref{sec:constructs}.}
In this section, we discuss the necessary theoretical background before presenting a toy example to illustrate our approach in Section \ref{sec:toyexample} and eventually proposing our theoretical framework in Section \ref{sec:framework}.

\subsection{Programming Language Theory} 
\label{ssec:proglangtheory}


We must have a universal representation of algorithms so that our framework can present a consistent representation to end users for interpretation.
For this we turn to control flow graphs.
The control flow graph (CFG) is a common tool in computer science.
CFGs are visual representations of computer program executions~\citep{allenControlFlowAnalysis1970,mccabeComplexityMeasure1976b}.\footnote{A control flow graph is a ``directed graph in which the nodes represent basic blocks, sequences of program instructions having one entry point (the first instruction executed) and one exit point (the last instruction executed), and the edges represent control flow paths'' \citep{allenControlFlowAnalysis1970}.}
CFGs are used to visualize the possible paths 
through a program's execution for tasks such as code debugging \citep{allenProgramDataFlow1976}, 
compiler optimization \citep{shiversControlflowAnalysisHigherorder1991}, and program tracing for automatic 
differentiation \citep{baydinAutomaticDifferentiationMachine2017}.

CFGs can include a variety of \textit{control structures} as nodes in the graph, such as sequences, \textsc{goto} statements, and others. 
For our purposes we would like to restrict the possible types of nodes in the graph to avoid overwhelming the users.
The \textit{structured program theorem} from computer science states 
that every algorithm can be represented as a control flow 
graph built with three types of control structures: 
\textit{sequences}, where one \textit{block} follows another; 
\textit{selections}, where a boolean expression determines the subsequent 
sequence; and \textit{iterations}, where a sequence is repeated while some 
boolean holds true \citep{bohmFlowDiagramsTuring1966a}.
The proof shows that complex control structures (e.g., recursion) can be decomposed to a combination of the above basic control structures.
Therefore, any algorithm for solving a computable function, including complex machine learning models, recommender systems, and other algorithms involved in business decisions, can be represented as a control flow graph. 

Algorithms for solving problems have existed for millenia \citep{barbinHistoryAlgorithmsPebble2012,cookeHistoryMathematicsBrief2011}, and are typically (informally) defined as ``procedure[s] for solving a mathematical problem$\dots$in a finite number of steps that frequently involves repetition of an operation.''\footnote{https://www.merriam-webster.com/dictionary/algorithm}
A more technical definition of an algorithm is any computable function, as per the Church-Turing thesis \citep{churchNoteEntscheidungsproblem1936,turingComputableNumbersApplication1936}. 
These computable functions are functions that, given any input, halt eventually and output the expected function output for the given input \citep{turingComputableNumbersApplication1936,agarwalLearnabilityWihComputable2020}.
This definition encompasses any algorithm that would be developed and applied in a production environment and is, therefore, suitable for our purposes.

The structured program theorem and the Church-Turing thesis give us a foundation for our framework. 
They provide two desirable properties (namely, \textit{universality} and \textit{computability}) that we can leverage for our construction of a measurement for algorithmic interpretability.
We have a precise definition of algorithms that applies to their standard use-case, and we have a representation (CFGs) that can apply to any algorithm that might be used in practice, from regression models to neural networks and complex pipelines.

\subsection{Cognitive Load Theory}
\label{ssec:theory-clt} 

Cognitive load theory (CLT), an instructional design theory in cognitive science, builds upon the characteristics of human cognitive architecture and develops guidelines for instructional design \citep{swellerCognitiveLoadTheory1994a}.
Human cognitive architecture involves limited-capacity working memory and effectively unlimited long-term memory 
\citep{baddeleyWorkingMemory1983,swellerPsychologyLearningMotivation2003,cowanWorkingMemoryCapacity2005}.

Working memory can only process a few elements at any time and may become overloaded when more than a few elements are processed simultaneously, leading to cognitive overflow.
In contrast, long-term memory can hold knowledge in the form of schemas, which are knowledge structures used to store and organize information \citep{paasCognitiveLoadTheory2003}.
The human schematic knowledge base in long-term memory represents the major critical factor influencing how we learn new information 
\citep{kalyugaCriticalThinkingHigher2014}.
A schema, regardless of its complexity, is processed as a single entity in working memory.
Thanks to schema construction, although the number of elements that can be processed in working memory is limited, the amount of information that can be processed in working memory is unlimited.
For example, simple operations used in algorithms such as addition and multiplication are constructed schemas available in long-term memory for most users, which require minimal conscious effort.
Furthermore, lower-level schemas can be incorporated into higher-level schemas and consequently further reduce working memory load.
For example, users with high algorithmic literacy may have the dot product schema, which is a higher-level schema compared to the addition schema and the multiplication schema. Users with low algorithmic literacy may not have acquired the dot product schema yet and, thus, have to rely on addition and multiplication (lower-level schemas) to learn dot product. When trying to make sense of new information (e.g., a new algorithm), users typically resort to the highest-order schemas available to reduce working memory load \citep{swellerCognitiveArchitectureInstructional1998a}.


According to CLT, there are three sources of cognitive load: extraneous, germane, and intrinsic cognitive load \citep{swellerCognitiveLoadTheory1994a,paasCognitiveLoadTheory2003}.
Extraneous cognitive load refers to the unnecessary demand on working memory capacity often caused by ineffective instructional procedures.
In contrast, germane cognitive load refers to the necessary demand on working memory capacity.
CLT further provides instructional design guidelines to eliminate extraneous cognitive load.
The focus of this paper is intrinsic cognitive load, i.e., the demand on working memory capacity that is intrinsic to the material being learned and determined by its information structure.

In this paper, we measure intrinsic algorithmic interpretability by measuring the intrinsic cognitive load of an algorithm, i.e., how hard it is for users to understand the algorithm.
According to CLT, intrinsic cognitive load is determined by both the nature of the materials being learned and the learner's expertise.
Furthermore, intrinsic cognitive load can be characterized in terms of element interactivity \citep{merrienboerCognitiveLoadTheory2005}.
Intrinsic cognitive load is higher if the number of interacting elements is higher because more elements have to be processed simultaneously to understand the material fully.\footnote{Throughout the rest of this paper, we will refer to 
intrinsic cognitive load as simply ``cognitive load.''}
Understanding a new algorithm is a complex task with high element interactivity.
Due to working memory limitations, users cannot assimilate all of the elements if all of the interacting elements are presented simultaneously.
Instead, users rely on gradually building increasingly complex knowledge structures in long-term memory to manage cognitive load.

Our framework considers the three determinants of intrinsic load including number of elements, interactivity of these elements, and prior knowledge of the learner.
Since we do not explicitly model the learning process of users, we do not consider the determinants for germane and extraneous loads, such as instructional approach and learning style.



\section{Illustrative Toy Example: Revenue Calculation}
\label{sec:toyexample}

\begin{figure}[h!]
	\scriptsize
	\begin{subfigure}{\textwidth}
		\centering
	    \includegraphics[width=0.45\columnwidth]{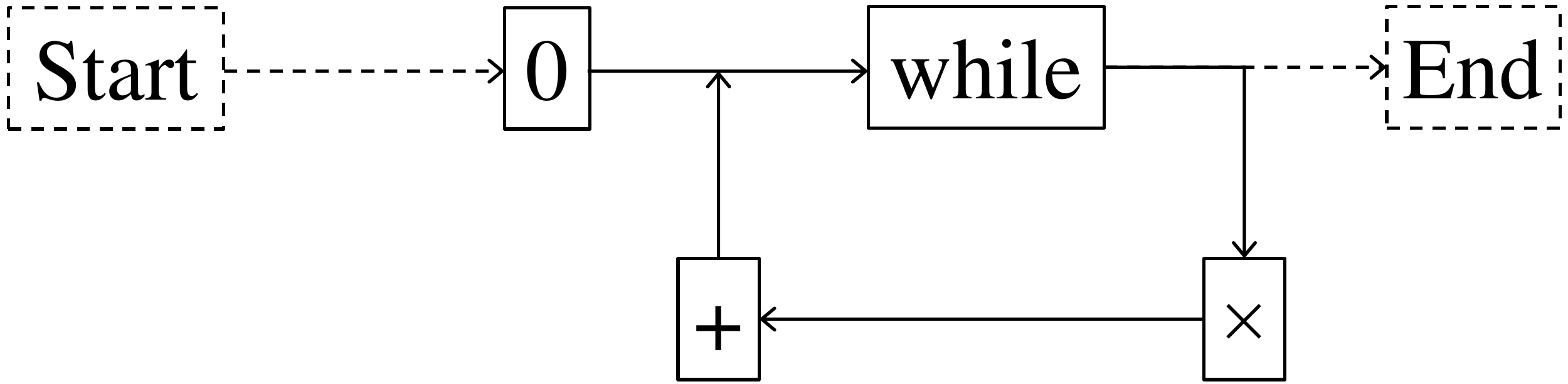}
		\newsubcap{Task 1 control flow graph}
		\label{fig:task1a}
	\end{subfigure}
	\begin{subfigure}{.5\textwidth}
	    \vspace{.2in}
		\centering
	    \includegraphics[width=0.9\columnwidth]{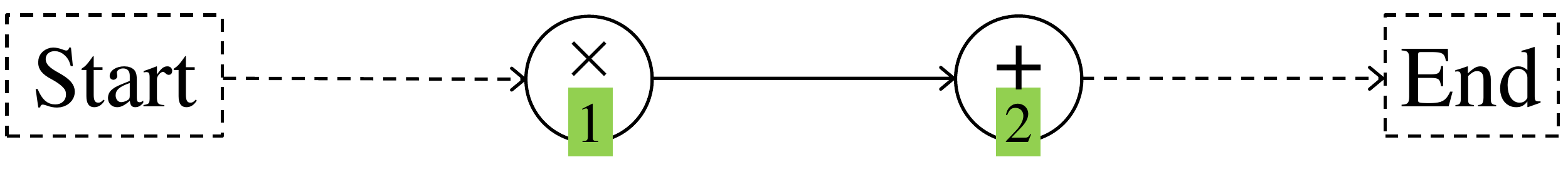}
		\newsubcap{Task 1 operation context graph for users \\[-.1in] with low algorithmic literacy\\[-.1in]}
		\label{fig:task1b}
	\end{subfigure}
	\begin{subfigure}{.5\textwidth}
	    \vspace{.2in}
		\centering
	    \includegraphics[width=0.9\columnwidth]{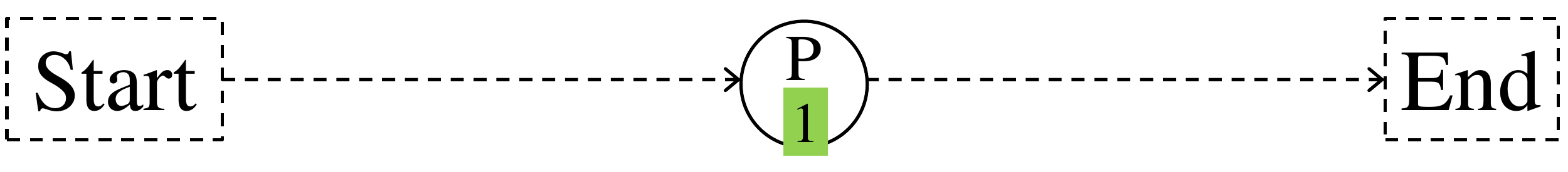}
		\newsubcap{Task 1 operation context graph for users \\[-.1in] with high algorithmic literacy\\[-.1in]}
		\label{fig:task1c}
	\end{subfigure}
	\caption{Task 1: Total revenue of all items}
	\label{fig:task1}
\end{figure}

\begin{figure}[h!]
	\scriptsize
	\begin{subfigure}{\textwidth}
		\centering
	    \includegraphics[width=0.45\columnwidth]{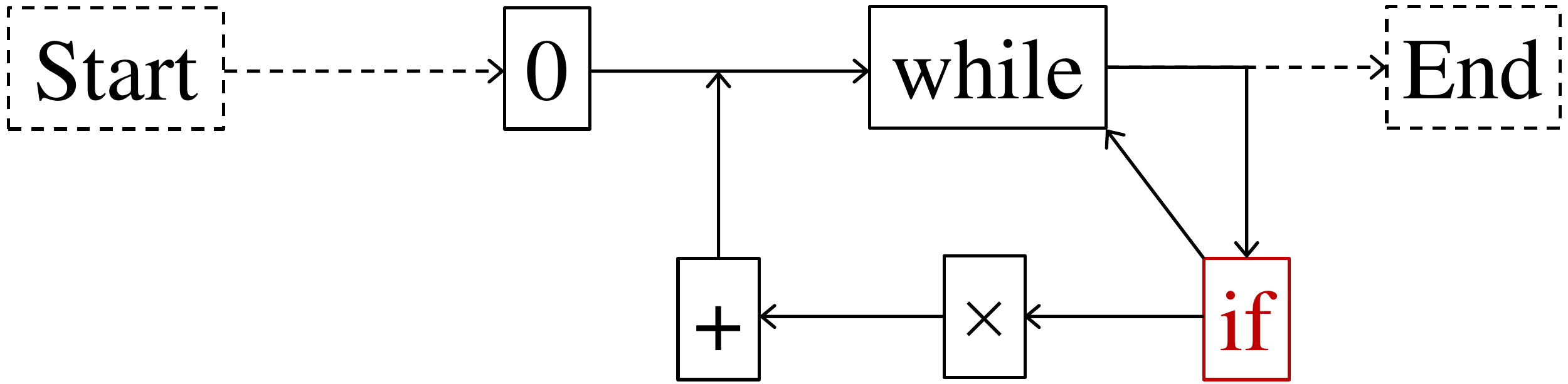}
		\newsubcap{Task 2 control flow graph}
		\label{fig:task2a}
	\end{subfigure}
	\begin{subfigure}{.5\textwidth}
	    \vspace{.2in}
		\centering
	    \includegraphics[width=0.9\columnwidth]{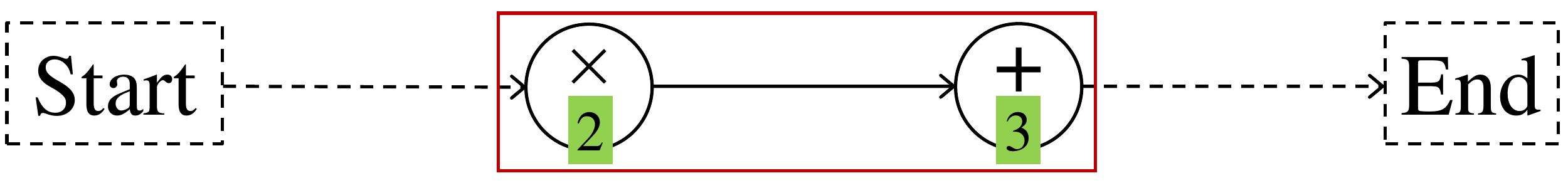}
		\newsubcap{Task 2 operation context graph for users \\[-.1in] with low algorithmic literacy\\[-.1in]}
		\label{fig:task2b}
	\end{subfigure}
	\begin{subfigure}{.5\textwidth}
	    \vspace{.2in}
		\centering
	    \includegraphics[width=0.9\columnwidth]{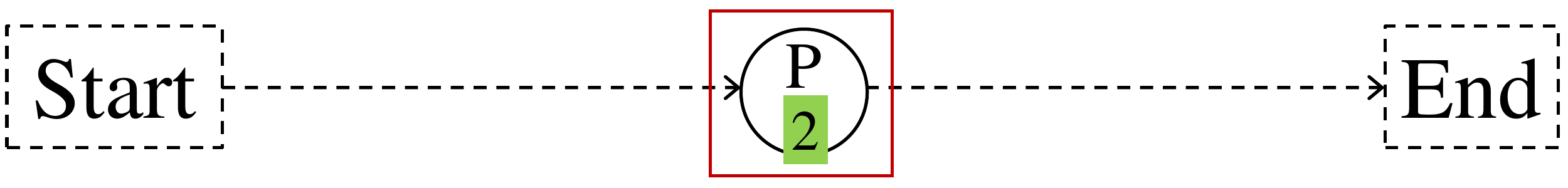}
		\newsubcap{Task 2 operation context graph for users \\[-.1in] with high algorithmic literacy\\[-.1in]}
		\label{fig:task2c}
	\end{subfigure}
	\caption{Task 2: Total revenue of items with price higher than \$10}
	\label{fig:task2}
\end{figure}

\begin{figure}[h!]
	\scriptsize
	\begin{subfigure}{\textwidth}
		\centering
	    \includegraphics[width=0.45\columnwidth]{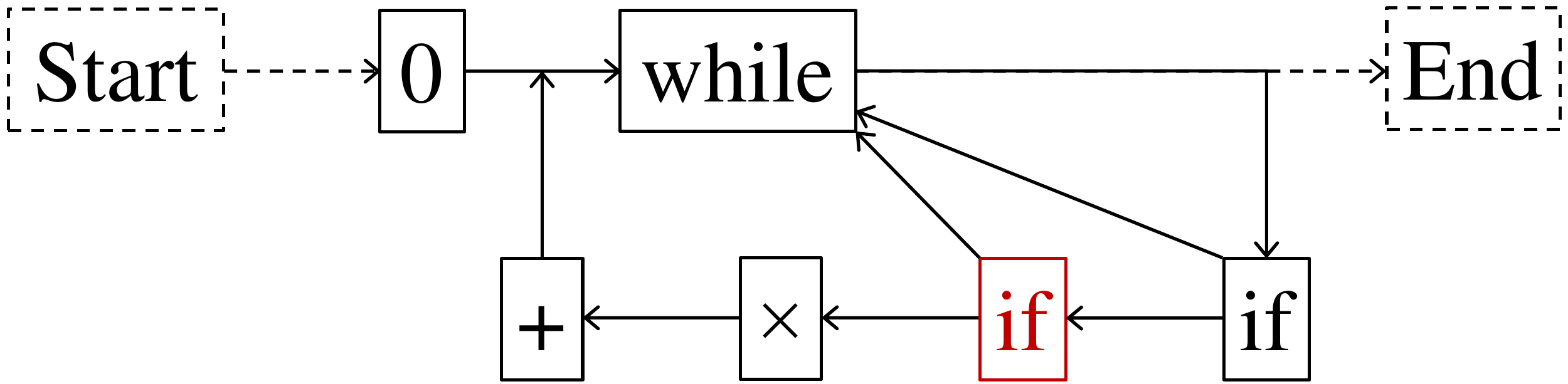}
		\newsubcap{Task 3 control flow graph}
		\label{fig:task3a}
	\end{subfigure}
	\begin{subfigure}{.5\textwidth}
	    \vspace{.2in}
		\centering
	    \includegraphics[width=0.9\columnwidth]{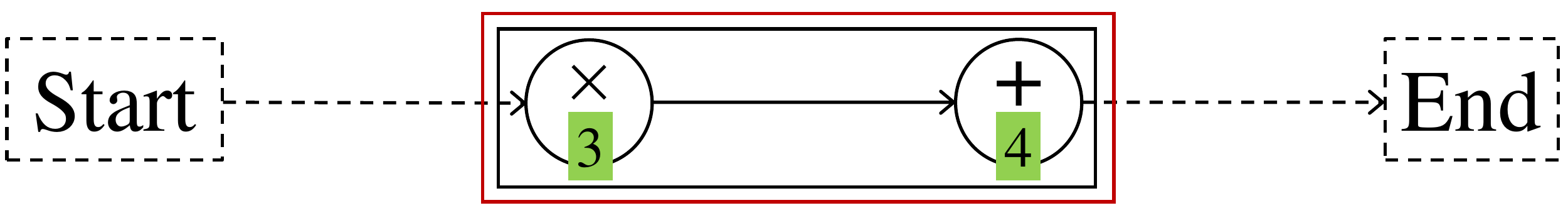}
		\newsubcap{Task 3 operation context graph for users \\[-.1in] with low algorithmic literacy\\[-.1in]}
		\label{fig:task3b}
	\end{subfigure}
	\begin{subfigure}{.5\textwidth}
	    \vspace{.2in}
		\centering
	    \includegraphics[width=0.9\columnwidth]{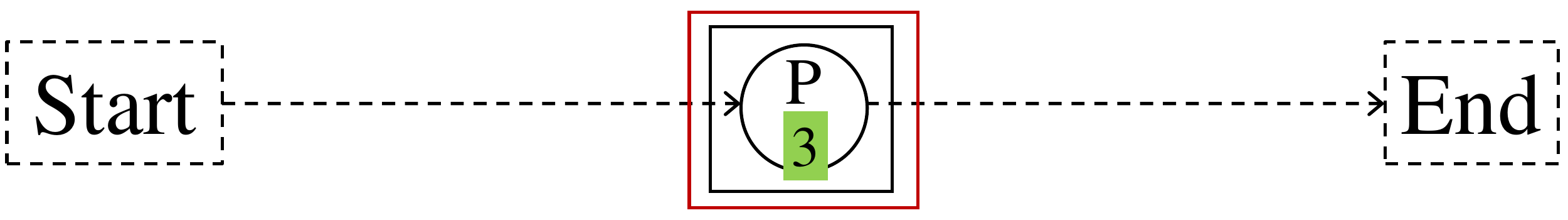}
		\newsubcap{Task 3 operation context graph for users \\[-.1in] with high algorithmic literacy\\[-.1in]}
		\label{fig:task3c}
	\end{subfigure}
	\caption{Task 3: Total revenue of items with price higher than \$10 and quantity sold higher than 100}
	\label{fig:task3}
\end{figure}

Before formally presenting our theoretical framework of measuring algorithmic interpretability, in  this section we discuss a simple toy example of revenue calculation to illustrate the main ideas in the framework. 
Given the price and quantity sold for a list of items, we would like to compute the total revenue of these items.
Specifically, we consider three tasks in this example.
In Task 1, as indicated in Figure \ref{fig:task1}, the total revenue of all items in the list is computed. 
In Task 2, as indicated in Figure \ref{fig:task2}, the total revenue of items with a price higher than $\$10$ is computed.
In Task 3, as indicated in Figure \ref{fig:task3}, the total revenue of items with a price higher than $\$10$ and a quantity sold higher than $100$ is computed.
Next, we elaborate on 
the detailed process of deriving the cognitive complexity scores in these three tasks.

\textbf{Task 1: Total revenue of all items.} Figure \ref{fig:task1a} shows the control flow graph of Task 1, which starts with initializing a variable and then iteratively adding to the variable the product of price and quantity for each item.
Figures \ref{fig:task1b} and \ref{fig:task1c} show the operation context graphs of Task 1 for users with low and high algorithmic literacy respectively. 
In Task 1, addition and multiplication are relevant schemas available for users with low algorithmic literacy and thus are the appropriate abstraction level in Figure \ref{fig:task1b}.
For users with high algorithmic literacy, they have dot product (a higher-level schema) available in their schematic knowledge base. Thus, dot product is the appropriate abstraction level in Figure \ref{fig:task1c}.

It is important to note that the variable initialization step and the while loop in the control flow graph do not show up in the operation context graph because, unlike computers, humans process information through concepts such as addition and multiplication, instead of going through a loop.\footnote{For example, 1000 additional iterations would add more computational complexity for computers. However, the cognitive complexity for humans remains the same.}
Additionally, since the concept of all (i.e., all items in the list) is an automated schema for the users, there is no specific context for the user to process in Task 1.
Therefore, the number of interacting elements (the key determinant of cognitive load) is 1 for the multiplication operation (the focal operation itself only) and 2 for the addition operation (the parent operation and the focal operation).
Since the increase in the number of interacting elements results in a nonlinear increase in the cognitive load imposed on working memory, we use an exponential function to capture such nonlinear growth patterns.\footnote{Our framework can be extended/adapted to other nonlinear functional forms.} 
Consequently, the resulting cognitive complexity score is $e+e^2$ for users with lower algorithmic literacy and $e$ for users with higher algorithmic literacy.

\textbf{Task 2: Total revenue of items with price higher than 
\$10.} As shown in Figure \ref{fig:task2a}, different from Task 1, the selection criterion of items with a price higher than $\$10$ is a context-specific condition in Task 2.
This additional condition requires extra effort for users to process.
Such additional conditions are extracted as contexts in our framework (indicated as rectangles in operation-context graphs), which also impose extra cognitive load and, thus, add to the final cognitive complexity score.
Specifically, the resulting cognitive complexity score is $e^2 + e^3$ for users with lower algorithmic literacy (Figure \ref{fig:task2b}) and $e^2$ for users with higher algorithmic literacy (Figure \ref{fig:task2c}).

\textbf{Task 3: Total revenue of items with price higher than \$10 and quantity sold higher than 100.}
In Task 3, as shown in Figure \ref{fig:task3a}, we have an additional piece of contextual information to consider.
Two criteria must be met instead of one.
This second criterion further increases cognitive load and the cognitive complexity score, which is indicated in Figures \ref{fig:task3b} and \ref{fig:task3c} by adding another context rectangle to the graphs.
Specifically, the resulting cognitive complexity score is $e^3 + e^4$ for users with lower algorithmic literacy (Figure \ref{fig:task3b}) and $e^3$ for users with higher algorithmic literacy (Figure \ref{fig:task3c}).

\section{A Framework for Measuring Intrinsic Algorithmic Interpretability}
\label{sec:framework}
\begin{figure*}[t]
	\centering 
	\includegraphics[width=1\textwidth]{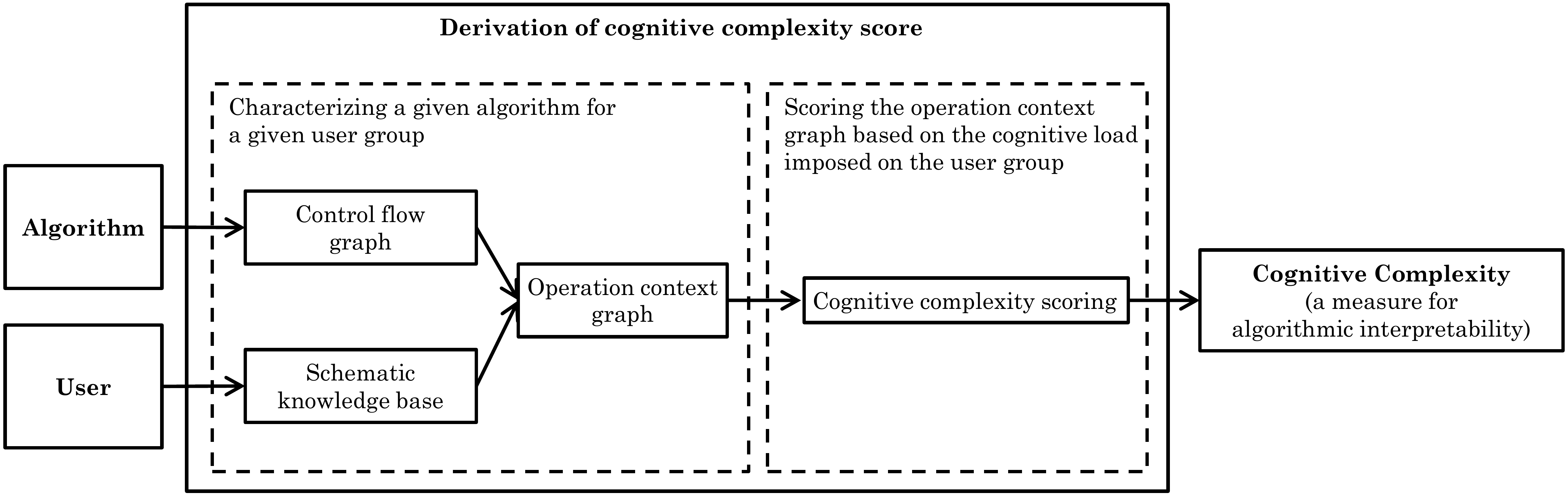}
	\caption{Research framework of measuring algorithmic interpretability}
	\label{fig:framework}
\end{figure*}

In this section, we propose a human-learning-oriented theoretical framework for measuring algorithmic interpretability (Figure \ref{fig:framework}). 
Our framework consists of four major components: representing algorithms as control flow graphs (\S \ref{ssec:plt}); characterizing user expertise with schemas (\S \ref{ssec:clt}); constructing the operation context graphs (\S \ref{ssec:graphconstruction}); and scoring for cognitive complexity (\S \ref{ssec:scoring}).
Before we discuss these components in detail, we first introduce the necessary notation for consistent exposition of the framework.

\subsection{Notation}
For a given algorithm $a$ and a user group $u$, we aim to measure the interpretability of the algorithm to the user group through measuring  the cognitive complexity for the user group to learn the algorithm.
For a specific algorithm $a$, we use $CFG(a)$ to represent the control flow graph of $a$.
Let $U$ be the set of \textit{user groups}, where user group $u \in U$ is defined as the set of users that share the same level of algorithmic literacy (i.e., the same set of available schemas).
For a specific user group $u$, we use $S_u$ to represent the schematic knowledge base of $u$.
For a given algorithm $a$ and a user group $u$, we use $OCG(a,u)$ to represent the operation context graph of $a$ appropriate for user group $u$.

\subsection{Representing Algorithms as Control Flow Graphs}
\label{ssec:plt} 

Based on programming language theory, a given algorithm can be represented by a control flow graph (\S \ref{ssec:proglangtheory}).
And by the structured program theorem, we know that these graphs only include a small set of flow types; namely, sequences, selections, and iterations.
This result allows for a universal, graphical representation of algorithms that is simple in terms of allowable components.\footnote{Certain algorithms may have large, complex control flow graphs, but they \textit{can} still be represented as such.}
By restricting acceptable components, we can standardize how the graphs are generated, making downstream analyses consistent.

\subsection{Characterizing User Expertise with Schemas}
\label{ssec:clt}

Different users have different levels of expertise for understanding algorithms.
User expertise needs to be abstracted so that the measurement can be applied according to some pre-set criteria.
Consider the logistic regression classifier as an example.
A machine learning researcher can abstract more concepts related to the classifier to higher-order representations than a legislator because the 
researcher has higher \textit{expertise} than the legislator.\footnote{Expertise 
can take many forms. Later we explicitly discuss technical knowledge such as 
algorithmic literacy, but other forms of expertise are relevant as well and are 
context- and algorithm-dependent.}
Both are interested in the interpretability of the model for different reasons 
(e.g., research, policy).

To characterize expertise, we resort to schemas from cognitive load theory.
When an individual learns a new concept, that concept is internalized in long-term memory as a schema, which can be recalled and applied when learning new, complex concepts \citep{paasCognitiveLoadTheory2003}.
A schema, regardless of its complexity, is processed as a single entity in working memory.
Thanks to schema construction, although the number of elements that can be processed in working memory is limited,\footnote{According to Miller's law, the ``magic number'' of objects one can hold in working memory is $7 \pm 2$ \citep{millerMagicalNumberSeven1994}; the amount of information that a human can process in working memory is unlimited.}
through (often extensive) practice, humans can learn and understand more complex concepts by building on existing internalized schemas.
In the context of algorithms, every block in a control flow graph can be learned and internalized as a schema.


A user group's internalized schemas can be summarized in their schematic knowledge base.
The schematic knowledge base then determines if any transformations have to be made to the control flow graph.
For example, if a user has a schema for taking an average, a basic block for calculating an average can be abstracted away as a single operation (i.e., \textit{TakeAverage}) and will be processed as a single entity in working memory.
However, if the user does not have a schema for taking an average, the operation would be decomposed into its 
parts.
Each part would be considered a separate operation: one for the addition schema, connected to one for the division schema, resulting in a larger graph.
The schematic knowledge base impacts the level of abstraction involved in the transformation from control flow graph to \graphname~graph.

\subsection{Construction of Operation Context Graphs}
\label{ssec:graphconstruction}

With the necessary pieces in place, we now describe the process of transforming control flow graphs to account for human expertise and cognitive load restrictions. 
Control flow graphs provide a consistent, universal representation of algorithms where subgraphs in a user's internalized schemas in long-term memory can be abstracted away and replaced with a single operation node in the operation context graph.

The structured program theorem allows for sequences, selections, and iterations as control structures.
Selections are involved in extracting contexts; they involve the evaluation of boolean conditions to determine which block to execute next.
The blocks can be transformed into operations according to acquired schemas.
Once the appropriate operations are identified, the additional cognitive load of tracking the boolean condition is represented by context.
Iterations over a full list do not affect context. 
However, if there are conditionals embedded in the list (e.g., only certain elements), then that selection is taken into account when extracting context.

Operations and contexts give us the framework's building blocks.
To generate a graph appropriate for a given user level, we start at the fully decomposed level, where it is assumed that the entire algorithm has been decomposed according to the structured program theorem so that only the most basic operations are included in the graph. 
Each subgraph is a possible decomposition of an internalized schema in the user's schematic knowledge base.
If so, that subgraph can be abstracted away and replaced in the graph by the higher-level schema.
The process continues until each block in the graph is a schema in long-term memory, and no subgraphs are decompositions of higher-level internalized schemas; then any selections are accounted for with contexts. 
This process generates a concrete, visual representation of an appropriate algorithm, given the schemas that a user has internalized in long-term memory.
It also takes into account the cognitive costs of tracking contexts. 

For a fully decomposed control flow graph representation of an algorithm and a data table of the schematic knowledge base, an appropriate graph can be constructed by 
abstracting subsections of the graph based on a user group's schematic knowledge base until each block is an operation in the schema set, and each selection has been extracted into context.

\subsection{Cognitive Complexity Scoring} 
\label{ssec:scoring}

We have shown that it is possible to generate a universal, human-oriented representation of any algorithm.
We next address the question of quantifying the interpretability of 
the operation 
context graphs for measuring algorithmic interpretability. 
Understanding a new algorithm is a complex task with high element interactivity.
Intrinsic cognitive load (hereafter ``cognitive load'') is higher if the number 
of interacting elements is higher because more elements have to be processed 
simultaneously to understand the material fully.
Due to working memory limitations, users first learn isolated elements with 
a partial understanding of the material and then learn interacting elements to 
understand the material fully \citep{merrienboerCognitiveLoadTheory2005}.

Based on the \graphname~graph, the \textit{cognitive 
complexity}~score for the algorithm can be computed as:
\vspace{-.2in}
\begin{align*}
	CC(a,u) &=  \sum_{n \in OCG(a,u)} e ^ {CL(n)} \\
	CL(n) &= C(n) + P(n) + 1 
\end{align*}
Here, $CC(a, u)$ is the cognitive complexity score of algorithm $a$'s representation appropriate for user group $u$.
$C(n)$ is the contextual state of node $n$, $P(n)$ is the number of parent nodes for $n$, and $1$ is the operational cost of the node itself.
$CL(n)$ is the congnitive load of node $n$, which tells us how many interacting elements must be considered at one time in working memory for a given operation.
The \textit{cognitive complexity} associated with increasing interacting elements is exponential, which can be extended/adapted to other nonlinear functional forms.
For an algorithm, the \scorename~is then the sum of the \scorename~scores of all operations in the graph.



For an algorithm and a group of users, the \scorename~score is an inverse measure of interpretability. 
The higher the score, the more difficult it is for users to interpret the algorithm.
Cognitive complexity captures two critical elements of intrinsic interpretability: algorithm complexity and prior human knowledge.
If someone wants to understand a particular algorithm, they cannot simply take it all in at once.
An algorithm consists of many steps, each of which must be processed to go from input to output.
Each operation involves interacting elements of parent nodes and contexts, which affect the cost of processing the node in working memory.
In addition, a block in an algorithm can only be understood if the user has an internalized schema for it.
If not, the block must be decomposed until the blocks are part of the user's schemas in long-term memory.

In Section \ref{sec:properties}, we will investigate the properties of the measurement framework and their implications. In Section \ref{sec:recsys}, we will discuss a real-life business example with recommender system algorithms~\citep{adomaviciusNextGenerationRecommender2005a,adomaviciusContextawareRecommenderSystems2011} as an application of the proposed measurement framework.


\section{Framework Properties and Implications}
\label{sec:properties}

The measurement framework and the corresponding \scorename~score have four desirable properties: universality, computability, uniqueness, and monotonicity.
Two properties (universality and computability) follow from our theoretical grounding, while two (uniqueness and montonicity) are new results from our formulation.
Proofs for all four properties are provided in Appendix \ref{sec:proofs}.

\begin{property}[Universality]
    \label{rmk:universality}
    For any algorithm $a$ and user group $u$, the proposed measurement framework is applicable to the $(a,u)$ pair.
\end{property}

\begin{property}[Computability]
    \label{rmk:computability}
    For any algorithm $a$ and user group $u$, the proposed measurement framework is machine-computable.
\end{property}

\begin{property}[Uniqueness]
    \label{rmk:uniqueness}
    For a given algorithm $a$ and a given user group $u$, the proposed measurement framework generates a unique cognitive complexity score $CC(a,u)$.
\end{property}

\begin{property}[Monotonicity]
	\label{rmk:montonicity}
	For a given algorithm $a$, and two user groups $u_{1}$ and $u_2$, if user group $u_1$ has lower algorithmic literacy than user group $u_2$, then the cognitive complexity score is higher for $u_1$ than $u_2$, (i.e., $\text{if } S_{u_1} \subset S_{u_2}, \text{then } CC(a,u_1) \geq CC(a,u_2))$.
\end{property}

With these properties, our framework can be used as a general measure of the cognitive complexity of algorithms. 
These properties reduce ambiguity and enforce consistency in how algorithms are represented, transformed, and scored.

The framework is universal and can be applied to any algorithm.
For any algorithm, a user either does or does not understand each ``step.''
The steps are themselves algorithms that can be decomposed into step-by-step 
procedures. 
This universality applies to simple algorithms such as those for taking an 
average and algorithms for training and deploying complex neural network 
models. 

Because generating the operation context graph is computable, the process of creating an operation-context graph and generating a cognitive complexity score can be automated. 
This automation is much less costly than user studies.
Cognitive complexity scores can be automatically generated for a large number of algorithms so that they can be compared simultaneously without the need for large amounts of human response data.

For a given algorithm and user group pairing, there is precisely one way to generate a representation. 
Only certain operations are available to the user at a certain level of 
expertise in a schematic knowledge base. 
Therefore, all of the individual components in the~\graphname~graph are fixed.
For the same reason, the \scorename~score is also unique for the algorithm and user level.
This gives consistency to the framework.

Monotonicity is an important measurement property for ordinal comparisons \citep{freemanOrderbasedStatisticsMonotonicity1986}.
In particular, when measuring a latent construct via observed data (in this case, algorithms and schemas are observed, and cognitive complexity is latent), monotonicity ensures that as one observed trait changes, the latent trait changes in a consistent direction.
For a given algorithm, it should always be the case that a group with more relevant expertise can more readily understand and interpret a particular algorithm.

\section{Real-life Business Example: Recommender Systems}
\label{sec:recsys}

To illustrate the framework, we walk through an example using two algorithms 
from Recommender Systems 
\citep[RecSys,][]{adomaviciusNextGenerationRecommender2005a,adomaviciusContextawareRecommenderSystems2011,ekstrandCollaborativeFilteringRecommender2011,kluverRatingbasedCollaborativeFiltering2018,lee2021product}.
RecSys models have seen wide adoption in recommending items to shoppers, movies to
viewers, and songs to listeners, among many other use cases.
We follow the steps outlined in our framework (Figure \ref{fig:framework}) to go from problem formulation through to cognitive complexity scores for each algorithm under consideration.

\subsection{Problem Formulation}

For a set of users $U$, a set of items $I$, and a (potentially sparse) matrix of user-item ratings $R$, a given user (e.g., user $u$ or $v$) and a given item (e.g., item $i$ or $j$), the goal is to predict a rating score
for a specific user to a specific item:

\begin{equation}
	\hat{S}_{u, i} = RecSys(R, u, i)
\end{equation}

where $RecSys$ is an algorithm for calculating the
predicted rating of $u$ for $i$.
Estimated scores can inform item-ranking decisions, e.g., which products or movies to display to the user.

\subsection{Algorithm Selection}

We consider two standard algorithms for RecSys tasks: the
\useritembias~(or personalized mean) algorithm (\uib) and the
\useruserknn~(\uuknn) algorithm
\citep{ekstrandCollaborativeFilteringRecommender2011,kluverRatingbasedCollaborativeFiltering2018}.
It is commonly accepted that the user item bias algorithm is ``more interpretable'' than the KNN algorithm.
Our goal is to validate this claim with our framework.

\uib~is a common baseline RecSys model due to its
simplicity and is often used for setting a performance lower bound
\citep{kluverRatingbasedCollaborativeFiltering2018}.
The predicted rating for a particular user and item is the sum of a global average rating, an item bias, and a
user bias:
\reqnomode

\begin{align}
	\label{eq:uibstart}
	\mu &= \frac{\sum_{r_{ui} \in R} r_{ui}}{\vert R \vert } \\
	\label{eq:ubi_itembias}
	b_i &= \frac{\sum_{u \in U_i} (r_{ui} - \mu)}{\vert U_i \vert}
	\\
	\label{eq:ubi_userbias}
	b_u &= \frac{\sum_{i \in I_u} (r_{ui} - b_i - \mu)}{\vert I_u
			\vert} \\
	\label{eq:uibend}
	\hat{S}_{u, i} &= \mu + b_i + b_u
\end{align}
For \uuknn, user $u$'s rating for item $i$ is estimated as the weighted average of $u$'s $k$-most similar users' (\textit{neighbors}') ratings for $i$, according to some pre-defined similarity metric.
The critical component to \uuknn~is identifying neighbors for some user $u$.
For our example, we use Pearson's $r$ as the similarity metric \citep[Equation \ref{eqn:knnLowSim},][]{herlockerEmpiricalAnalysisDesign2002,kluverRatingbasedCollaborativeFiltering2018}.
After calculating and sorting similarity scores for all possible $v$'s, the top-$k$ are selected, and the predicted score is the weighted average of the neighbors' scores for item $i$. 
Weights are based on the neighbors' similarity to $u$, so more similar users 
will have a more significant influence on the predicted rating (Equation \ref{eqn:knnLowS}). 

\begin{align}
	\label{eqn:knnLowMu}
	\mu_u &= \frac{\sum_{r_{ui} \in R_u} r_{ui}}{\vert R_u \vert } \\
	\label{eqn:knnLowSim}
	sim(u, v) &= \frac{\sum_i (r_{ui} - \mu_u)(r_{vi} - \mu_v)}{\sqrt{\sum_i (r_{ui} - \mu_u)^2} \times \sqrt{\sum_i (r_{vi} - \mu_v)^2}} \\
	\label{eqn:knnLowN}
			N_{u} &= \{v_i \in V_{sorted} |  1 \leq i \leq k   \} \\ 
	\label{eqn:knnLowS}
	\hat{S}_{u, i} &= \frac{\sum_{v \in N_{u}} sim(u, v) \times r_{vi}}{\sum_{v \in N_{u}} \vert sim(u,v) \vert } 
\end{align}
\label{fig:knnlowmath}

Certain operations can be combined to present an appropriate representation for 
a high algorithmic literacy user.
Here we can treat $u$'s and $v$'s ratings for items that they have both rated as vectors ($\mathbf{r}_u$ and $\mathbf{r}_v$).
$sim(u, v)$~can be re-written as the dot product of $u$'s and $v$'s adjusted ratings vectors divided by the product of the L2-norms of $u$'s and $v$'s adjusted ratings vectors (Equation \ref{eqn:knnHighSim}).
For both algorithms, there are extensions and modifications for improving 
performance (e.g., the selection of similarity score for \uuknn).
In our framework, these are simply additional operations that can be added to 
the graph for each algorithm, affecting the \scorename~score.

\begin{align}
	\label{eqn:knnHighMu}
	\mu_u &= \frac{\sum_{r_{ui} \in R_u} r_{ui}}{\vert R_u \vert } \\
	\tilde{\mathbf{r}}_u &= \mathbf{r}_u - \mu_u\mathbf{I} \\
	\label{eqn:knnHighSim}
	sim(u, v) &= \frac{\tilde{\mathbf{r}}_u \cdot \tilde{\mathbf{r}}_v}{\vert \vert \tilde{\mathbf{r}}_u \vert \vert_2  \times \vert \vert \tilde{\mathbf{r}}_v \vert \vert_2}\\
	\label{eqn:knnHighN}
			N_{u} &= \{v_i \in V_{sorted} |  1 \leq i \leq k   \} \\ 
	\label{eqn:knnHighS}
	\hat{S}_{u, i} &= \frac{\sum_{v \in N_{u}} sim(u, v) \times r_{vi}}{\sum_{v \in N_{u}} \vert sim(u,v) \vert } 
\end{align}
\label{fig:knnhighmath}

\subsection{Capturing User Expertise with Schemas}
Building schematic knowledge bases for users requires making certain assumptions about what operations are internalized as schemas in long-term memory.
For this example we consider a relatively coarse delineation of user 
expertise: low and high expertise levels in terms of users' algorithmic literacy.
We set these two algorithmic-literacy levels using testing information for a U.S. high school equivalency examination (HiSET),\footnote{https://hiset.ets.org/about/content} which does not include vector and matrix multiplication, and the GRE mathematics subject test used in graduate school admissions,\footnote{https://www.ets.org/gre/subject/about/content/mathematics/} which does.
We can assume that vector and matrix multiplication operations, which are relevant for our analyses, are not part of the expected schemas of low expertise individuals, but are part of the expected schemas of high expertise individuals.

\subsection{Incorporating Schemas to a Control Flow Graph to Generate an \Graphname~Graph}
\label{sssec:baselinegraph}

\begin{figure}[h!]
	\scriptsize
	\begin{subfigure}{0.6\textwidth}
		\centering
	    \includegraphics[width=\columnwidth]{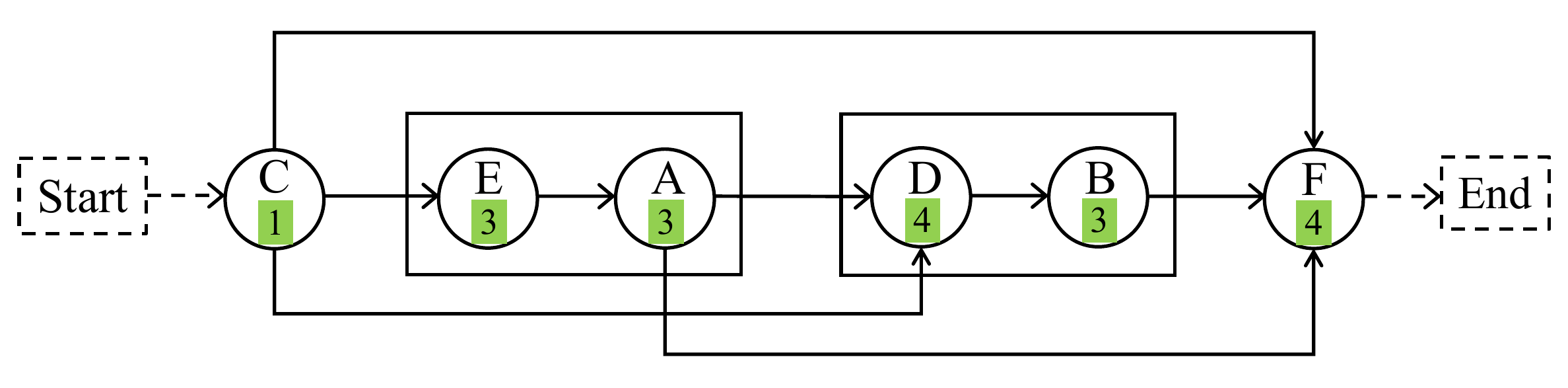}
	\end{subfigure}
	\begin{subfigure}{0.4\textwidth}
		\centering
		\begin{tabular}{c}
		     Legend for Figure \ref{fig:uib-all}\\
		    \begin{tabular}{ll}
		    	\toprule
			    Node & Description \\
			    \midrule 
			    A&Calculate average: $b_i$\\
			    B& Calculate average: $b_u$\\
			    C&Calculate global average: $\mu$\\
			    D&Subtract $\mu$ and $b_i$ from user ratings\\
			    E&Subtract $\mu$ from item ratings\\
			    F&Sum global average, user bias, and \\
			    &item bias values\\
			    \bottomrule
			    \\
		    \end{tabular}
		\end{tabular}
	\end{subfigure}
	\caption{Operation context graph of the \useritembias~algorithm for users at two algorithmic literacy levels}
	\label{fig:uib-all}
\end{figure}

\begin{figure}[h!]
	\begin{subfigure}{\textwidth}
		\centering
		\includegraphics[width=.65\columnwidth]{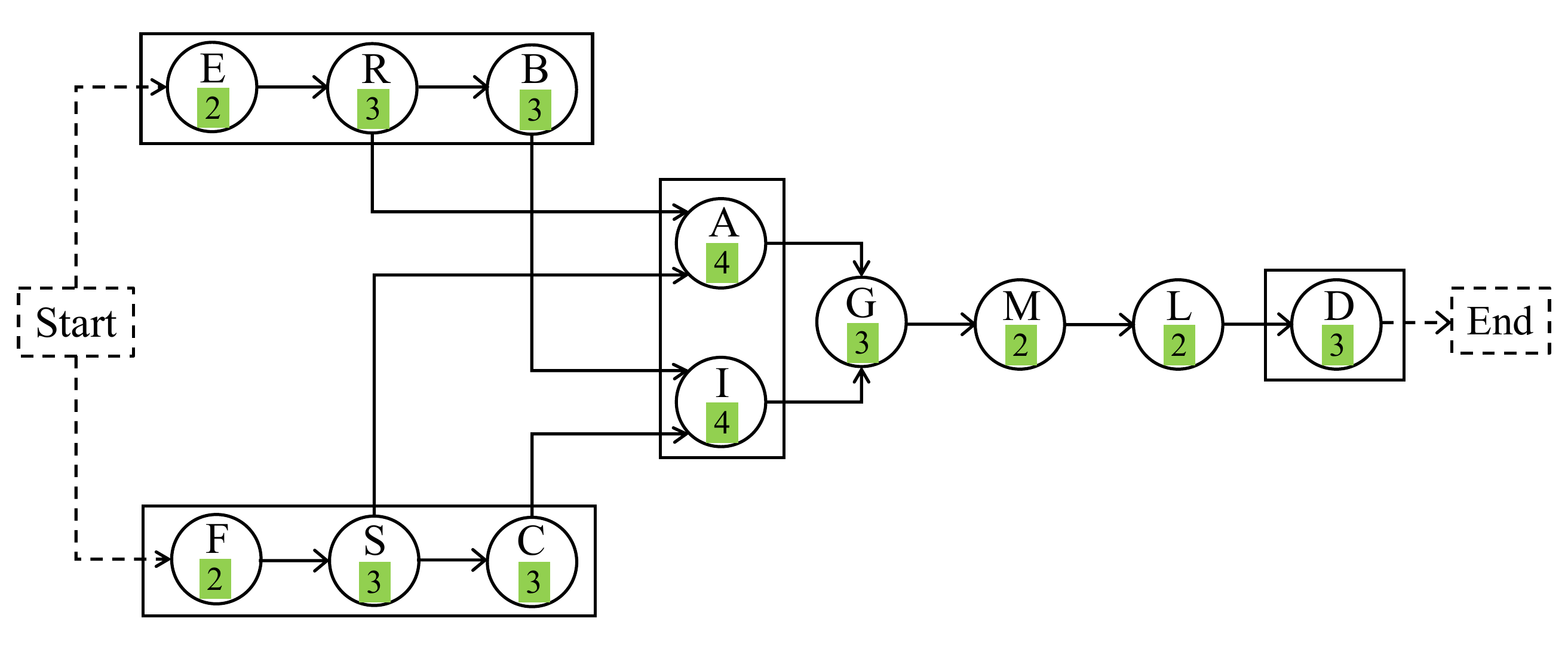}
        \newsubcap{Operation context graph of \useruserknn~algorithm for users with high algorithmic literacy}
		\label{fig:knn-high-skill}
	\end{subfigure}
	\begin{subfigure}{\textwidth}
	    \vspace{.2in}
		\centering 
	    \includegraphics[width=.8\columnwidth]{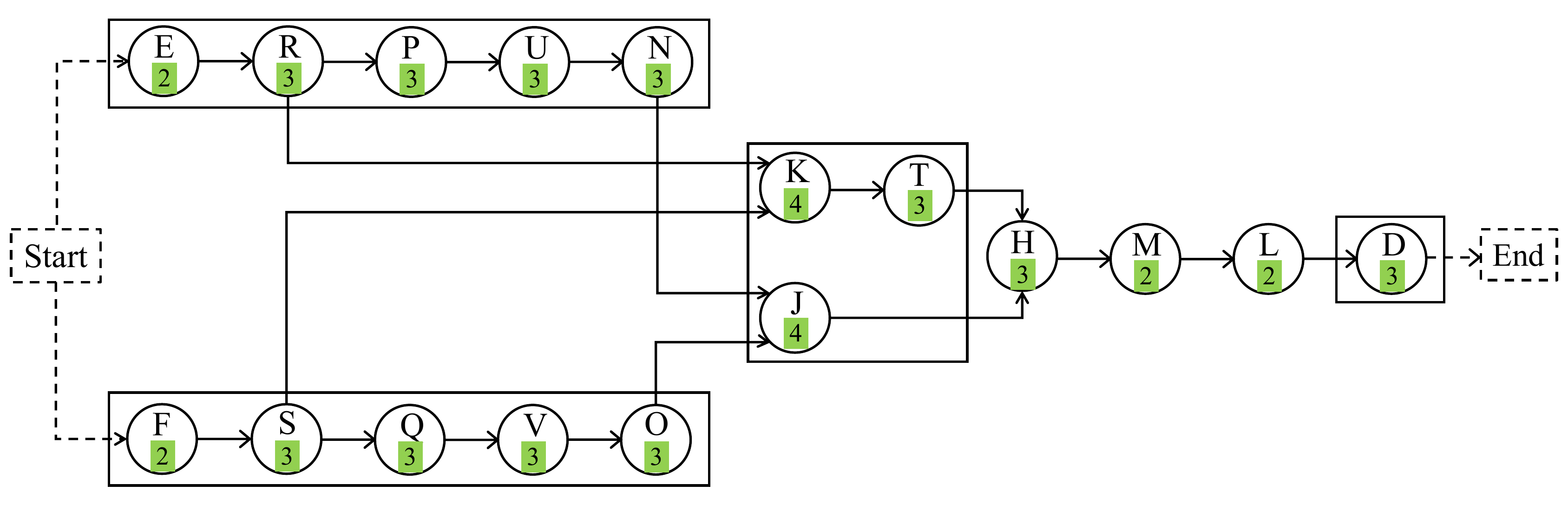}
		\newsubcap{Operation context graph of \useruserknn~algorithm for users with low algorithmic literacy}
		\label{fig:knn-low-skill}
	\end{subfigure}
	\scriptsize 
	\begin{subfigure}{\textwidth}
		\vspace{.2in}
		\begin{tabular}{c}
		    Legend for Figure \ref{fig:knn-high-skill} and Figure \ref{fig:knn-low-skill}\\
		    \\
			\begin{tabular}{lp{4.5cm}c}
				\toprule
				Node & Description & \shortstack{Algorithmic\\ Literacy Level}\\
				\midrule 
				A&Calculate the dot product of the two adjusted ratings vectors&High\\
				B&Calculate the L2 norm of $u$'s adjusted ratings&High\\
				C&Calculate the L2 norm of $v$'s adjusted ratings&High\\
				D&Calculate the weighted average of the neighbors' ratings for $i$, using the similarity scores as the weights&Low\\
				E& Calculate $u$'s average rating&Low\\
				F&Calculate $v$'s average rating&Low\\
				G&Divide the dot product value by the product of L2 norms&High\\
				H&Divide the numerator by the denominator for similarity score&Low\\
				I&Multiply the L2 norm values together&High\\
				J&Multiply the square rooted values together&Low\\
				\bottomrule
				\\
			\end{tabular}
			\quad
			\begin{tabular}{lp{5cm}c}
				\toprule
				Node & Description & \shortstack{Algorithmic \\Literacy Level} \\
				\midrule 
				K&Multiply $u$'s and $v$'s adjusted ratings together for each item&Low\\
				L&Select the top-$k$ users based on similarity as $u$'s neighbors&Low\\
				M&Sort the similarity scores for all users&Low\\
				N& Square root $u$'s sum&Low\\
				O& Square root $v$'s sum&Low\\
				P&Square $u$'s adjusted ratings&Low\\
				Q&Square $v$'s adjusted ratings&Low\\
				R&Subtract $u$'s average rating from $u$'s rating for each item&Low\\ 
				S&Subtract $v$'s average rating from $v$'s rating for each item&Low\\
				T& Sum the products of adjusted ratings&Low\\
				U& Sum $u$'s squared adjusted ratings&Low\\
				V& Sum $v$'s squared adjusted ratings&Low\\
				\\
				\\
				\\
				\bottomrule
				\\
			\end{tabular}
		\end{tabular} 
		\centering 
	\end{subfigure}
	\caption{Operation context graph of the \useruserknn~algorithm for users at two levels}
	\label{fig:knn-all}
\end{figure}

Figures \ref{fig:uib-all} and \ref{fig:knn-all} show the operation context graphs for the two algorithms.
The \graphname~graph for \uib~is appropriate for both high and low skill users based on our schematic knowledge base (Figure \ref{fig:uib-all}).
This algorithm is simpler than the \uuknn~algorithm, therefore it is possible to construct a clear representation for both high and low algorithmic literacy users using high-school level mathematical operations.

For \uuknn, the representation for low algorithmic literacy users is complex (Figure \ref{fig:knn-all}).
Incorporating the vector and matrix multiplication abstractions for high 
algorithmic literacy users allows for a more interpretable graph.
However, these schemas are not available to the low algorithmic literacy users, who must instead use the decomposed graph to understand the algorithm.

Context is key to our framework as it marks selections and iterations where additional information must be dealt with. 
In the mathematical formulation, context is implicitly defined according to notation.
For example, calculating item bias in \uib~requires subtracting the global average from each rating in $U_i$ (Equation \ref{eq:ubi_itembias}).
In the graphical representation context is explicit.
A rectangle surrounds nodes that share context.
%
For \uib~there are two local contexts to consider: calculating item bias $b_i$ by iterating over all users who have rated item $i$ and calculating user bias $b_u$ by iterating over all items rated by user $u$.
For \uuknn~there are four contexts for iterations involving different subsets of data involved in certain operations.

After selections have been replaced with contexts, the \graphname~graph is complete.
From here \scorename~can be calculated (\S \ref{ssec:scoring}).
The levels of complexity in the graphs are reflected in the \scorename~scores (Table \ref{tab:interpretability-scores}).

\subsection{Using Cognitive Costs to Derive a Cognitive Complexity Score}
\label{sec:discussion}

To empirically assess the tradeoffs of cognitive complexity and other performance metrics we ran both RecSys algorithms on the MovieLens 100K data set \citep{harperMovielensDatasetsHistory2015}.
Table \ref{tab:interpretability-scores} includes cognitive complexity scores as well as computational complexity and predictive performance metrics.
Table \ref{tab:interpretability-scores} confirms what is anecdotally evident: 
the user-item bias algorithm is more interpretable (i.e., lower score) than the 
user-user KNN algorithm; also, a given algorithm is more interpretable (lower 
score) for high algorithmic literacy users than for low algorithmic literacy 
users.
None of the other metrics consider both the algorithm and the user as part of the scoring process.

\begin{table}[h]
	\centering \footnotesize
	\caption{Cognitive complexity scores as compared to computational complexity and performance metrics.}
	\label{tab:interpretability-scores}
	\begin{tabular}{lcc}
		\toprule
		Metric & \capitalizetitle{\useritembias} & \capitalizetitle{\useruserknn} \\
		\midrule
		Cognitive Complexity (Low Expertise) & $2e^4 + 3e^3 + e = 172.17$ & $2e^4 + 11e^3 + 4e^2 = 359.69$ \\
		Cognitive Complexity (High Expertise) & $2e^4 + 3e^3 + e = 172.17$ & $2e^4 + 6e^3 + 4e^2 = 259.27$\\
		Runtime Complexity, Actual (Upper Bound) & $2,700$ ($O(n + d))$& $1,705,492$ ($O(nd + n\log n)$) \\
		Space Complexity, Actual (Upper Bound) & $1,700,000$ ($O(nd))$& $1,701,700$ ($O(nd + n)$) \\
		Net Discounted Cumulative Gain (nDCG) & 0.073 & 0.093 \\
		F1 & 0.057& 0.064\\
		RMSE & 0.872& 0.791\\
		\bottomrule 
	\end{tabular} 
\end{table} 

\subsection{Tradeoffs between Cognitive Complexity and other Performance Metrics}
Managers must consider multiple metrics when making decisions about which algorithms to deploy. 
Table \ref{tab:interpretability-scores} shows the scores of all measures in the three key performance dimensions (algorithmic interpretability, computational complexity, and predictive performance). 
Among all measures in these three dimensions, user-item bias outperforms user-user $k$-nearest neighbors in terms of cognitive complexity and runtime complexity. 
The two algorithms have similar predictive performance although user-user $k$-nearest neighbors scores slightly better in F1 and Net Discounted Cumulative Gain (nDCG). 
Given the salient gain in interpretability, user-item bias may be a better option for managers.
As shown in this example, the measurement framework gives managers a quantitative metric of interpretability against which they can assess the performance of the candidate algorithms under consideration.

\section{Conclusion}
\label{sec:conclusion}

In this work we present a framework for \frameworkname. We build upon programming language theory and cognitive load theory to model algorithm intrinsic complexity and human cognitive constraints. In this section, we discuss our contributions to both research and practice.

\subsection{Contributions to Research}
\label{ssec:cResearch}
\textit{The first measurement framework for quantifying algorithmic interpretability, a new dimension for analyzing algorithms.}
In the 1960s, to assess the efficiency of running computer programs and storing data, scholars in the field of programming language theory recognized the need for studying the computational complexity of algorithms \citep{hartmanisComputationalComplexityAlgorithms1965,knuthSelectedPapersAnalysis2000}.
Researchers developed methods to ``make quantitative assessments of the goodness of various algorithms''~\citep{knuthSelectedPapersAnalysis2000}.
Today, in order to build trust, ensure fairness, and track accountability, there is a need to study a new dimension of goodness in algorithms -- algorithmic interpretability \citep{hardtEqualityOpportunitySupervised2016a,xiaoOpportunitiesChallengesDeveloping2018a,kearnsEthicalAlgorithmScience2019,mittelstadtExplainingExplanationsAI2019a,zhangEffectConfidenceExplanation2020}.

Just as scholars developed measurement frameworks for computational complexity to quantify how hard it is for \textit{computers} to \textit{compute} algorithms, in this study we are among the first to propose a measurement framework for algorithmic interpretability to quantify how hard it is for \textit{humans} to \textit{understand} algorithms. 
Based on these measurement frameworks, metrics such as runtime complexity and space complexity are derived for computational complexity; in this work, we derive the cognitive complexity score for algorithmic interpretability.

\paragraph{Foundation for future downstream research in quantitative algorithmic interpretability.}
Computational complexity frameworks served as the foundation for later research such as program optimization in software engineering~\citep{knuthArtComputerProgramming1997a,knuthSelectedPapersAnalysis2000}.
Our measurement framework and the corresponding cognitive complexity score serve as a theoretical foundation for emerging research areas in information systems such as algorithmic fairness, accountability, transparency, and ethics.
On the technical side, IS researchers develop and analyze interpretable algorithms~\citep[e.g.,][]{wangHybridPredictiveModels2021}.
This research stream can use cognitive complexity to quantify model interpretability, instead of intuition-driven proxies for quantifying interpretability.

More broadly, within the ESG-ICE (environmental, social, and governance-individual well-being, community welfare, and economic resilience) framework for using analytics to address societal challenges  proposed in \cite{ketterSpecialIssueEditorial2020}, effective measurement of algorithmic interpretability helps identify obscure components in analytics projects and subsequent effort can be exerted to improve transparency.
Inscrutability is identified as an important facet of AI~\citep{berenteManagingArtificialIntelligence2021}.
The formal measurement framework proposed in this paper serves as a theoretical foundation to quantify this important facet of AI.




\textit{Basis for improving the algorithm and the user learning process.}
Because we characterize expertise using specific operations, users with low expertise can see an appropriate representation of an algorithm and a list of operations associated with a high expertise representation. 
The framework takes into account what an individual has already learned when identifying appropriate operations.
Characterizing levels of expertise also provides a specific learning opportunity for users with low expertise. 
If they can understand those operations, they can understand a more compact, higher-level representation of the algorithm.
For our toy example, if a low expertise user wants to understand revenue calculation better, he or she could learn how to calculate dot products.
This framework provides a benefit to educators.
It offers a reliable, unique representation of an algorithm at a particular level of expertise to facilitate instruction.
As user expertise increases, the cognitive complexity of the algorithm monotonically decreases.

\textit{Potential implementation of a widely applicable and cost-effective measurement method.}
Because of the structured program theorem, the framework presented is general enough to account for any algorithm, even complex ones such as deep neural networks.
This universal property means that complex algorithms can be broken down into operations and contexts and represented as (complex) \graphname~graphs. 
It has long been assumed that such algorithms were less interpretable than simpler algorithms (or not interpretable at all).
Now such comparisons can be quantified and tested to illustrate how wide the intrinsic interpretability gap is.
This will inform discussions of algorithmic interpretability beyond notions of ``black-box'' algorithms and intuition-based distinctions between interpretable and not interpretable algorithms \citep{rudinStopExplainingBlack2019a,leavittFalsifiableInterpretabilityResearch2020a}.
The framework is also computable.
Implementations of the framework are cost-effective ways to automatically calculate algorithmic interpretability, as opposed to more costly user studies.

\subsection{Contributions to practice} 
\label{ssec:cPractice}

\Frameworkname~can benefit a variety of stakeholders.
Managers, users, policymakers, learners, and educators can use this framework to explicitly quantify intrinsic interpretability instead of relying on heuristics, intuition-driven categorizations, or post-hoc human assessments.

\paragraph{Managers:}
When selecting the best algorithm for a task at hand, managers have to balance several key tradeoffs, such as predictive performance, computational complexity, and interpretability. 
The \scorename~score quantifies how interpretable each candidate algorithm is to a 
specific user group. 
Considering all related user groups, managers can accurately assess the interpretability of the algorithms and then balance it against other  performance measures to make an informed decision.

\paragraph{Users:} 
With more algorithms being used in consumer products, external users such as suppliers and customers want to know how and why certain algorithmic decisions are being made.
Interpretability can also affect how much a user trusts a particular firm. 
Users can request an algorithmic representation at an appropriate level for them and compare firms in terms of how interpretable their algorithmic processes are.
Users may not want to see the operation context graphs for algorithms, although they are available, and may instead be simply interested in comparing cognitive complexity scores.

With algorithms being implemented in business processes, internal users will want a level of understanding regarding the new tools available in their workflows (e.g., a recruiter using an AI-powered resume screener).
An understanding of the algorithm will be useful for them in making sense of the decisions and providing feedback based on domain knowledge to make improvements to the algorithm.

\paragraph{Policymakers:} 
Recently passed laws that require a “right to explanation” from algorithmic decision systems do not specify the level of detail required for an explanation to suffice~\citep{goodmanEuropeanUnionRegulations2017a}.
Some understanding of how the algorithms generate outputs is needed.
Policymakers and regulators could set specific cognitive complexity guidelines for certain tasks, where an algorithm used in a particular context must have a score below a certain threshold.
For example, regulation could be passed stating that for user groups with a high school education, algorithms used for loan processing decisions cannot have a cognitive complexity score above a certain threshold.
This gives a concrete user group (and therefore schematic knowledge base) and specific score for firms to evaluate their models against.
This score can also serve as a proxy for cost, as a more complex algorithm may require hiring a subject matter expert to assess the algorithm's internal operations.\footnote{	
A quantitative interpretability score can inform discussions on algorithm 
accountability and ethical considerations 
\citep{abbasiMakeFairnessDesign2018,martinEthicalImplicationsAccountability2019a,berenteManagingArtificialIntelligence2021}. 
For certain high-stakes decision areas, one could consider an interpretability ceiling, where a deployed model cannot be too complex so that decisions made can be interpreted in the context of the underlying algorithm. 
Additionally, \scorename~scores can help policy makers to clarify requirements. For example, GDPR requires data controllers to provide the data subjects with meaningful information about the logic of model decision-making.}


\paragraph{Learners and Educators:}
Operation context graphs are a useful pedagogical device that implicitly encode a curriculum for those who would like to better understand particular algorithms.
Each operation node is something that can potentially be internalized as a schema. 
If a user wants to understand an algorithm better, then the \graphname~graph provides a clear roadmap for what to learn next.  

Our framework also identifies bottlenecks for learners.
Because of the cognitive cost of high element interactivity, certain nodes in an operation context graph are more cognitively demanding than others.
Educators can identify these complex sections of the graph and teach the schematic knowledge needed to abstract them to a higher-level schema.

This framework provides a much-needed theoretical grounding for measuring algorithmic interpretability.
Our hope is that this work acts as a call to the research community to motivate work on quantitative, theoretically grounded algorithmic interpretability and incorporate it into broader research in algorithmic fairness, accountability, transparency, and ethics.

\bibliographystyle{informs2014}
\bibliography{MAInt}
\begin{APPENDICES}

\newpage
\begin{landscape}
\section{Relevant Theoretical Concepts}
\label{sec:constructs}
\begin{table}[h!]
    \centering
    \caption{Key concepts in programming language theory (PLT) and cognitive load theory (CLT)}
    \label{tab:constructs}
    \scriptsize
    \begin{tabular}{lp{4.5cm}p{16cm}}
    \toprule
    Theory & Concept & Description \\
    \midrule
    PLT & Control flow & Control flow refers to the order in which program operations are executed by a computer.\\
    &Control flow graph & Graphical representation of the operations and flow of a program execution.\\
    &Sequence & A sequence is a control flow where one block follows another.\\
    &Selection &A selection is a control flow where the computer must evaluate a boolean expression before moving to the next block.\\
    &Iteration &An iteration is a control flow where the computer repeats a set of blocks until some boolean becomes true. \\
    &Structured program theorem & The structured program theorem states that every algorithm can be represented as a control flow graph using only sequences, selections, and operations to define the control flow.\\
    CLT&Working memory & Working memory has limited capacity and may become overloaded.\\
    &Long-term memory & Long-term memory can hold knowledge in the form of schemas and is effectively unlimited. \\
    &Schema & Schema is a knowledge structure used in long-term memory. Lower-level schemas can be incorporated into high-level schemas. Each schema, regardless of its complexity level, is processed as a single entity in working memory. \\
    &Human schematic knowledge base  & Human schematic knowledge base represents human expertise and is the major critical factor influencing how we learn new information. \\
    &Intrinsic cognitive load & Intrinsic cognitive load is determined by both the nature of the materials being learned and the learner's expertise. Intrinsic cognitive load can be characterized in terms of element interactivity. Intrinsic cognitive load is higher if the number of interacting elements is higher because more elements have to be processed simultaneously to fully understand the material.\\
    &Germane cognitive load & Germane cognitive load refers to the necessary demand on working memory capacity.\\
    &Extraneous cognitive load & Extraneous cognitive load refers to the unnecessary demand on working memory capacity often caused by ineffective instructional procedures.\\
    
    \bottomrule
    \end{tabular}
\end{table}

\end{landscape}
\newpage 

\section{Proofs of Framework Properties}
\label{sec:proofs}
\setcounter{property}{0}

\begin{property}[Universality]
    \label{rmk:universality}
    For any algorithm $a$ and user group $u$, the proposed measurement framework is applicable to the $(a,u)$ pair.
\end{property}

Proof: By the structured program theorem, any algorithm can be represented as a control flow graph \citep{bohmFlowDiagramsTuring1966a}.
For a user group with no internalized schemas (or no relevant schemas), then the representation is a fully-decomposed version of the algorithm (i.e., machine level code).
For any schema that is added and relevant, it will reduce the graph size.
For a user group that has fully internalized the algorithm as a schema, the graph will be fully abstracted. $\blacksquare$

\begin{property}[Computability]
    \label{rmk:computability}
    For any algorithm $a$ and user group $u$, the proposed measurement framework is machine-computable.
\end{property}

Proof: Once $a$ and $u$ are known, then the process of constructing an operation-context graph is fully automated.
By the definition of computable functions, the conversion to operation-context graph can be computed as a sequence of steps.
Specifically, from the fully decomposed graph representation, 
the problem of identifying the appropriate schemas for abstraction reduces to a subgraph isomorphism problem, a well-studied problem in computer science \citep{cookComplexityTheoremProvingProcedures1971}.\footnote{Graph isomorphism is an \textsc{NP-Hard} problem in computer science, for which a number of solvers exist \citep[e.g.,][]{solnonAlldifferentbasedFilteringSubgraph2010,carlettiChallengingTimeComplexity2017}.}
 $\blacksquare$

\begin{property}[Uniqueness]
    \label{rmk:uniqueness}
    For a given algorithm $a$ and a given user group $u$, the proposed measurement framework generates a unique cognitive complexity score $CC(a,u)$.
\end{property}

Proof:
Here we assume that $u$ will rely on the highest-level schemas available in the internalized knowledge base.
This is due to the fact that the users will rely on the higher-level schemas when available and, therefore, will not have a situation where there is a conflict between two possible lower-level schemas that can replace a subgraph in $G$. $\blacksquare$

\begin{property}[Monotonicity]
	\label{rmk:montonicity}
	For a given algorithm $a$, and two user groups $u_{1}$ and $u_2$, if user group $u_1$ has lower algorithmic literacy than user group $u_2$, then the cognitive complexity score is higher for $u_1$ than $u_2$, (i.e., $\text{if } S_{u_1} \subset S_{u_2}, \text{then } CC(a,u_1) \geq CC(a,u_2))$.\footnote{This applies to cases where expertise is single-dimensional, i.e., a lay person and physician or a low algorithmic literacy and high algorithmic literacy individual. In this case $S_{u_{1}}$ is a superset of $S_{u_{2}}$.}
\end{property}

Proof: If $S_{u_{1}} = S_{u_{2}}$, then the result is trivial. 
In the case where $S_{u_{1}}$ is a proper superset of $S_{u_{2}}$, then the starting point of our operation-context graph is $OCG(a,  u_{1})$.
Therefore, if we use any new operations in $S_{u_{2}}$ to change the graph, it will be to further abstract away subsections.
This reduction will reduce the number of nodes in the graph and, therefore, the cognitive complexity of the graph. $\blacksquare$

\end{APPENDICES}
\end{document}